\documentclass{sig-alternate}

\usepackage{xspace}
\usepackage{numprint}
\usepackage{subcaption}
\usepackage{algorithm,algorithmic}
\usepackage{booktabs}
\usepackage{mathtools}
\usepackage{url}
\usepackage{hyperref}
\usepackage{xspace}
\usepackage{color}
\usepackage{amssymb}
\usepackage{latexsym}

\newcommand{\ccol}[1]{\multicolumn{1}{c}{#1}}

\newcommand{\argmax}{\operatornamewithlimits{argmax}}

\newcommand{\mad}{\textsc{mad}\xspace}
\newcommand{\lp}{\textsc{lp}\xspace}

\newcommand{\tweetLID}{\texttt{tweetLID}\xspace}

\newcommand{\en}{\textsc{en}\xspace}
\newcommand{\es}{\textsc{es}\xspace}
\newcommand{\pt}{\textsc{pt}\xspace}
\newcommand{\ca}{\textsc{ca}\xspace}
\newcommand{\gl}{\textsc{gl}\xspace}
\newcommand{\eu}{\textsc{eu}\xspace}
\newcommand{\amb}{\textsc{amb}\xspace}
\newcommand{\und}{\textsc{und}\xspace}

\newcommand{\urls}{\textsc{url}s\xspace}
\newcommand{\scikit}{\texttt{scikit-learn}\xspace}

\begin{document}
%
% --- Author Metadata here ---
%\CopyrightYear{2007} % Allows default copyright year (20XX) to be over-ridden - IF NEED BE.
%\crdata{0-12345-67-8/90/01}  % Allows default copyright data (0-89791-88-6/97/05) to be over-ridden - IF NEED BE.
% --- End of Author Metadata ---

\title{Discriminating between similar languages in Twitter using label propagation}
\numberofauthors{2}
\author{
% 1st. author
\alignauthor Will Radford\\
 \affaddr{Xerox Research Centre Europe}\\
 \affaddr{6 chemin de Maupertuis}\\
 \affaddr{38240 Meylan, France}\\
 \email{will.radford@xrce.xerox.com}
% 2nd. author
\alignauthor Matthias Gall\'e\\
 \affaddr{Xerox Research Centre Europe}\\
 \affaddr{6 chemin de Maupertuis}\\
 \affaddr{38240 Meylan, France}\\
 \email{matthias.galle@xrce.xerox.com}
}

%\alignauthor
%Anonymous\\
%       \affaddr{Anonymous}\\
%       \affaddr{Anonymous}\\
%       \affaddr{Anonymous}\\
%       \email{Anonymous}
%% 2nd. author
%\alignauthor
%Anonymous\\
%       \affaddr{Anonymous}\\
%       \affaddr{Anonymous}\\
%       \affaddr{Anonymous}\\
%       \email{Anonymous}

\maketitle
\begin{abstract}
Identifying the language of social media messages is an important first step in linguistic processing.
Existing models for Twitter focus on content analysis, which is successful for dissimilar language pairs.
We propose a label propagation approach that takes the social graph of tweet authors into account as well as content to better tease apart similar languages.
This results in state-of-the-art shared task performance of 76.63\%, 1.4\% higher than the top system.
\end{abstract}

%A category including the fourth, optional field follows...



\section{Introduction}
Language identification is a crucial first step in textual data processing and is considered feasible over formal texts \cite{McNamee2005}.
The task is harder for social media (e.g. Twitter) where text is less formal, noisier and can be written in wide range of languages.
%Approaches that use textual context features have been shown to work well \cite{Schulze2000}.
We focus on identifying \emph{similar} languages, where surface-level content alone may not be sufficient.
Our approach combines a content model with evidence propagated over the social network of the authors.
For example, a user well-connected to users posting in a language is more likely to post in that language.
Our system scores 76.63\%, 1.4\% higher than the top submission to the \tweetLID workshop.\footnote{\url{http://komunitatea.elhuyar.org/tweetlid}}

\section{Background}
\label{sect:stoa}
Traditional language identification compares a document with a language fingerprint built from n-gram bag-of-words (character or word level). %~\cite{Schulze2000}. % function words~\cite{Grefenstette1995} or a graph of character n-grams~\cite{Tromp2011,Vogel2012}.
Tweets carry additional metadata useful for identifying language, such as geolocation \cite{Carter2013}, username \cite{Bergsma2012,Carter2013} and \urls mentioned in the tweet \cite{Bergsma2012}. % and Goldszmidt2013, Graham2014

Other methods expand beyond the tweet itself to use a histogram of previously predicted languages, those of users \texttt{@}-mentioned and lexical content of other tweets in a discussion \cite{Carter2013}.
Discriminating between similar languages was the focus of the VarDial workshop \cite{zampieri-EtAl:2014:VarDial}, and most submissions used content analysis.
%In general, ensemble systems have been found to perform well, either using a ``late-fusion'' method \cite{Carter2013} or a weighted combination \cite{Lui2014}.
These methods make limited use of the social context in which the authors are tweeting -- our research question is ``Can we identify the language of a tweet using the social graph of the tweeter?''.

Label propagation approaches \cite{ZhuCMU02} are powerful techniques for semi-supervised learning where the domain can naturally be described using an undirected graph.
Each node contains a probability distribution over labels, which may be empty for unlabelled nodes, and these labels are propagated over the graph in an iterative fashion.
Modified Adsorption (\mad)\cite{Talukdar:2009:NRA:1617459.1617489}, is an extension that allows more control of the random walk through the graph.
Applications of \lp and \mad are varied, including video recommendation \cite{Baluja:2008:VSD:1367497.1367618} and sentiment analysis over Twitter \cite{speriosu-EtAl:2011:UNSUP}. % relation extraction \cite{chen-EtAl:2006:HLT-NAACL06-Short} and class-instance mining \cite{talukdar-pereira:2010:ACL}.
%The ability to propagate or smooth labels over a graph has proven useful in cases where supervised data is limited.
%\mad has been used for time-limited part-of-speech annotation, iteratively annotating tokens or sentences, then propagating labels to unlabelled instances \cite{garrette-baldridge:2013:NAACL-HLT}.
%Another graph propagation algorithm was successfully used to train a semi-supervised conditional random field for part-of-speech tagging \cite{subramanya-petrov-pereira:2010:EMNLP}, allowing unlabelled instances to play a role in training.

\section{Method}
\label{sect:method}

Our method predicts the language $\ell$ for a tweet $t$ by combining scores from a content model and a graph model that takes social context into account, as per Equation \ref{eq:hybrid}:

\vspace{-3mm}
\begin{equation}
  \textit{lang}(t) = \argmax_\ell \lambda_1 p(\ell | t, \theta_{\textit{content}}) + \lambda_2 p(\ell | t, \theta_{\textit{social}})
  \label{eq:hybrid}
\end{equation}
\vspace{-3mm}

Where $\theta_{\textit{content}}$ are the content model parameters, $\theta_{\textit{social}}$ the social model parameters.\footnote{We do not optimise $\lambda_1$ and $\lambda_2$, setting them to 0.5.}

\subsection{Content model}
%Although graph-based semi-supervised learning techniques allow us to combine evidence from a graph in a principled way, it is difficult to use them to predict labels for nodes that are not strongly connected to the graph.
%We attempt to solve the ``cold-start'' language identification problem using a supervised classifier that predicts one vs. all for a set of languages.
Our content model is a 1 vs. all $\ell_2$ regularised logistic regression model\footnote{We use \scikit: http://scikit-learn.org}
with character 2- to 5-grams features, not spanning over word boundaries.
The scores for a tweet are normalised to obtain a probability distribution.

\subsection{Social model}
\begin{figure}[t!]
  \centering
  \includegraphics[width=0.45\textwidth]{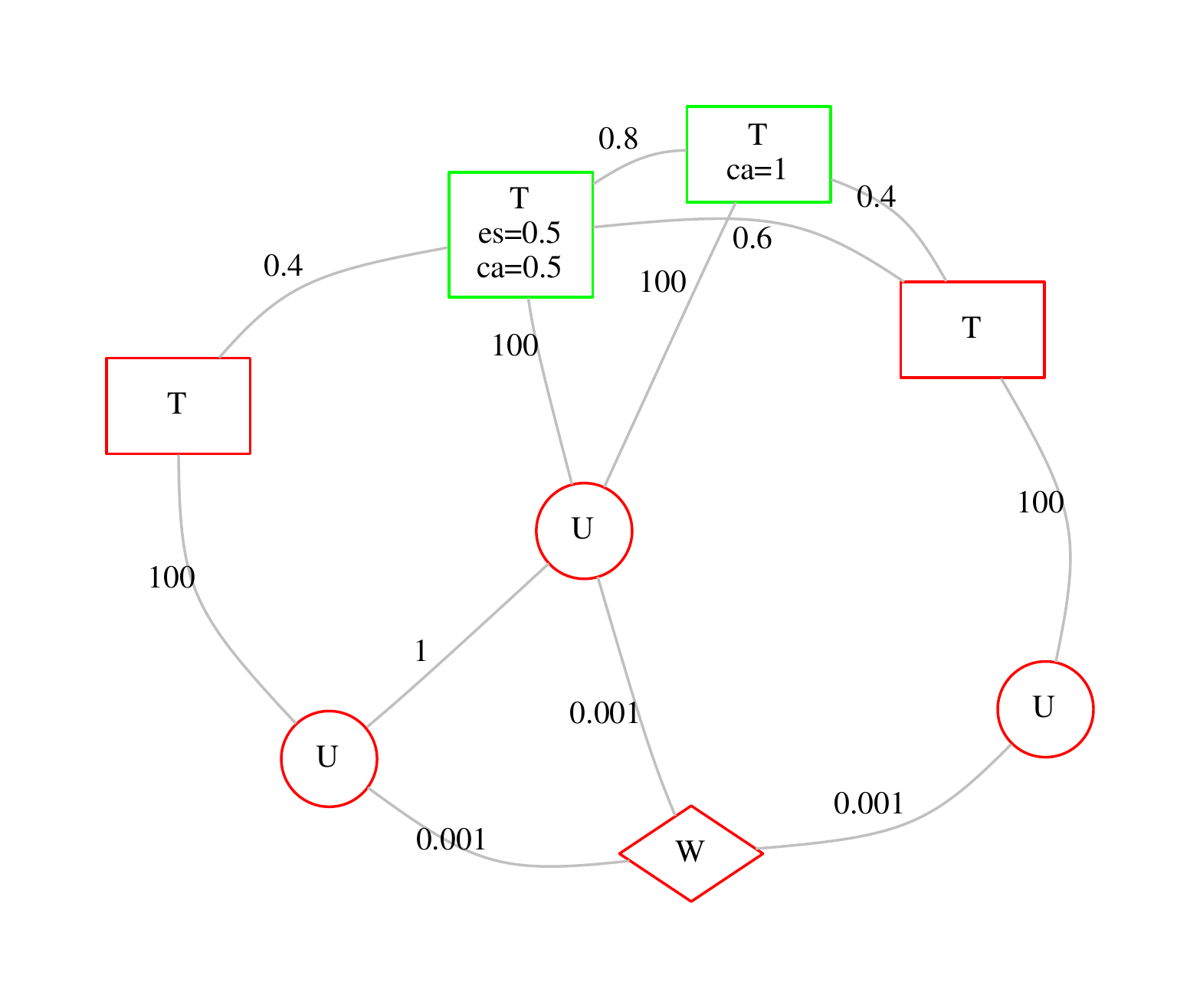}
  \vspace{-5mm}
  \caption{Graph topology. Rectangular nodes are tweets, circular nodes are users and the diamond represents the world. Some tweet nodes are \textcolor{green}{labelled} with an initial distribution over language labels and others are \textcolor{red}{unlabelled}.}
  \label{fig:graph}
\end{figure}

We use a graph to model the \emph{social media} context, relating tweets to one another, authors to tweets and other authors.
Figure \ref{fig:graph} shows the graph, composed of three types of nodes: tweets (T), users (U) and the ``world'' (W).
Edges are created between nodes and weighted as follows: \textbf{T-T} the unigram cosine similarity between tweets, \textbf{T-U} weighted 100 between a tweet and its author, \textbf{U-U} weighted 1 between two users in a ``follows'' relationship and \textbf{U-W} weighted 0.001 to ensure a connected graph for the \mad algorithm.

We create the graph using all data, and training set tweets have an initial language label distribution.\footnote{We assume a uniform distribution for \amb tweets.}
A na\"ive approach to building the tweet-tweet subgraph requires O($n^2$) comparisons, measuring the similarity of each tweet with all others.
Instead, we performed $k$-nearest-neighbour classification on all tweets, represented as a bag of unigrams, and compared each tweet and the top-$k$ neighbours.\footnote{We used \scikit with $k=0.25*ntweets$.}
We use Junto (\mad) \cite{Talukdar:2009:NRA:1617459.1617489} to propagate labels from labelled to unlabelled nodes.
Upon convergence, we renormalise label scores for initially unlabelled nodes to find the value of $\theta_{graph}$.

\section{Evaluation}
The \tweetLID workshop shared task requires systems to identify the language of tweets written in Spanish (\es), Portuguese (\pt), Catalan (\ca), English (\en), Galician (\gl) and Basque (\eu).
Some language pairs are similar (\es and \ca; \pt and \gl) and this poses a challenge to systems that rely on content features alone.
We use the supplied evaluation corpus, which has been manually labelled with six languages and evenly split into training and test collections.
% TODO {What about the 532 missing tweets?}
%532 tweets ($2.8\%$) of the tweets were not available any more when we downloaded them in Oct 2014.
%Results are reported on the remaining set.
We use the official evaluation script and report precision, recall and F-score, macro-averaged across languages.
This handles ambiguous tweets by permitting systems to return any of the annotated languages.
% TODO {Removed \urls?}
%The only pre-processing we performed was to remove \textsc{url}s.
%Each classifier (one per language) was run and it received the assigned label as long as the threshold was above $0.5$ (we did not try to optimize this threshold).
%The results with respect to the baseline (2-5 character $n$-grams) are shown in Table~\ref{table:res}.
Table~\ref{table:res} shows that using the content model alone is more effective for languages that are distinct in our set of languages (i.e. English and Basque).
For similar languages, adding the social model helps discriminate them (i.e. Spanish, Portuguese, Catalan and Galician), particularly those where a less-resourced language is similar to a more popular one.
Using the social graph almost doubles the F-score for undecided (\und) languages, either not in the set above or hard-to-identify, from 18.85\% to 34.95\%.
Macro-averaged, our system scores 76.63\%, higher than the best score in the competition: 75.2\%.

%Note that -- assuming the results extrapolate to the real test data -- the baseline would be somewhere in the middle of the leaderboard, while the hybrid model beats the reported best result ($75.2$)\footnote{\url{http://komunitatea.elhuyar.org/tweetlid/participation/#Results}}.

%\begin{table}[t!]
%  \centering
%  \begin{tabular}{lrr}
%    \toprule
%    Language & Train & Test \\
%    \midrule
%    es & 8,437 & 11,975 \\
%    pt & 2,102 &  1,957 \\
%    ca & 1,555 &  1,600 \\
%    en & 1,127 &  1,069 \\
%    gl &   748 &    607 \\
%    eu &   508 &    452 \\
%    und &  192 &    911 \\
%    \bottomrule
%  \end{tabular}
%  \caption{Data distribution.}
%  \label{tab:data}
%\end{table}

%\begin{table}
%\centering
%\begin{tabular}{||l|rrr|rrr||}
%\hline
%\hline
%	&\multicolumn{3}{c|}{CONTENT}	&	\multicolumn{3}{c||}{HYBRID} \\
%	& P 	& R 	& F 	& P 	& R 	& F  \\
%\hline
%ca	&95.14	&86.93	&90.82	&95.76	&90.62	&93.10\\
%gl	&66.76	&43.74	&52.65	&65.27	&51.04	&57.10\\
%en	&90.45	&76.10	&82.61	&91.12	&74.50	&81.92\\
%es	&91.45	&96.09	&93.71	&92.94	&95.79	&94.34\\
%eu	&93.25	&63.48	&75.39	&91.98	&66.06	&76.75\\
%pt	&93.46	&91.76	&92.58	&93.79	&94.56	&94.16\\
%amb	&100.00	&87.89	&93.48	&100.00	&91.91	&95.72\\
%und	&58.92	&20.51	&29.84	&67.91	&22.18	&32.72\\
%\hline
%avg	&86.18	&70.81	&\textbf{76.39}	&87.34	&73.33	&\textbf{78.23}\\
%
%\hline
%\hline
%\end{tabular}

\begin{table}
  \centering
  \begin{tabular}{lrrrrrr}
    \toprule
      &\multicolumn{3}{c}{Content}	&	\multicolumn{3}{c}{Content + Social} \\
      & \ccol{P} 	& \ccol{R} 	& \ccol{F} 	& \ccol{P} 	& \ccol{R} 	& \ccol{F}  \\
      \midrule
\es $\diamondsuit$     &92.64  &95.69  &94.14  &93.55  &95.89  &{\bf 94.70}\\
\pt $\spadesuit$     &89.81  &92.58  &91.17  &94.87  &92.52  &{\bf 93.68}\\
\ca $\diamondsuit$     &81.14  &87.19  &84.06  &85.22  &90.17  &{\bf 87.62}\\
\en      &77.42  &76.18  &{\bf 76.79}  &77.86  &70.53  &74.01\\
\gl $\spadesuit$    &56.93  &52.93  &54.85  &65.15  &50.35  &{\bf 56.80}\\
\eu      &92.41  &76.29  &{\bf 83.58}  &94.41  &68.01  &79.06\\

      \midrule
amb     &100.00 &89.56  &{\bf 94.49}  &100.00 &85.54  &92.21\\
und     &66.67  &10.98  &18.85  &45.06  &28.54  &{\bf 34.95}\\
\midrule
avg     &82.13  &72.67  &74.74  &82.01  &72.69  &{\bf 76.63}\\
    \bottomrule
  \end{tabular}

  \caption{Experimental results. $\diamondsuit/\spadesuit$ are similar pairs.}
  \label{table:res}
\end{table}

\section{Conclusion}
Our approach uses social information to help identify the language of tweets.
This shows state-of-the-art performance, especially when discriminating between similar languages.
A by-product of our approach is that users are assigned a language distribution, which may be useful for other tasks.

\bibliographystyle{abbrv}
\bibliography{biblio_db}  % sigproc.bib is the name of the Bibliography in this case
% ACM needs 'a single self-contained file'!
%
\end{document}